\documentclass[conference,letterpaper]{IEEEtran}
\IEEEoverridecommandlockouts

\usepackage{cite}
\usepackage{amsmath,amssymb,amsfonts}
\usepackage{algorithmic}
\usepackage{graphicx}
\usepackage{textcomp}
\usepackage{xcolor}
\usepackage{todonotes}
\usepackage{float}
\def\BibTeX{{\rm B\kern-.05em{\sc i\kern-.025em b}\kern-.08em
    T\kern-.1667em\lower.7ex\hbox{E}\kern-.125emX}}
\begin{document}

\title{End-to-end Generative Spatial-Temporal Ultrasonic Odometry and Mapping Framework}

\author{
    \IEEEauthorblockN{
        Fuhua Jia\textsuperscript{1}, 
        Xiaoying Yang\textsuperscript{2}, 
        Mengshen Yang\textsuperscript{1},
        Yang Li\textsuperscript{1}, 
        Hang Xu\textsuperscript{2}, 
        Adam Rushworth\textsuperscript{1},\\ 
        Salman Ijaz\textsuperscript{1}, 
        Boon Giin Lee\textsuperscript{2}, 
        Heng Yu\textsuperscript{2}, 
        Tianxiang Cui\textsuperscript{2}
    }
    \IEEEauthorblockA{
        \textsuperscript{1}Department of Mechanical, Materials and Manufacturing, University of Nottingham Ningbo China \\
        \textsuperscript{2}School of Computer Science, University of Nottingham Ningbo China\\
        % \textsuperscript{4}Nottingham Ningbo China Beacons of Excellence Research and Innovation Institute\\
        % Emails: fuhua.jia@nottingham.edu.cn, xiaoying.yang@nottingham.edu.cn
    }
}

\maketitle

\begin{abstract}
Performing simultaneous localization and mapping (SLAM) in low-visibility conditions, such as environments filled with smoke, dust and transparent objets, has long been a challenging task. Sensors like cameras and Light Detection and Ranging (LiDAR) are significantly limited under these conditions, whereas ultrasonic sensors offer a more robust alternative. However, the low angular resolution, slow update frequency, and limited detection accuracy of ultrasonic sensors present barriers for SLAM. In this work, we propose a novel end-to-end generative ultrasonic SLAM framework. This framework employs a sensor array with overlapping fields of view, leveraging the inherently low angular resolution of ultrasonic sensors to implicitly encode spatial features in conjunction with the robot's motion. Consecutive time frame data is processed through a sliding window mechanism to capture temporal features. The spatiotemporally encoded sensor data is passed through multiple modules to generate dense scan point clouds and robot pose transformations for map construction and odometry. The main contributions of this work include a novel ultrasonic sensor array that spatiotemporally encodes the surrounding environment, and an end-to-end generative SLAM framework that overcomes the inherent defects of ultrasonic sensors. Several real-world experiments demonstrate the feasibility and robustness of the proposed framework. 
\end{abstract}
%  The joint data containing spatial and temporal features is enhanced into dense scan through transformer. Then, the convolutional neural network (CNN) processes the spatiotemporal feature data and multi-frame scan to generate the pose transformation information at the current timestamp. By linearly superimposing the continuous scan with the pose transformation, the odometer and map can be constructed. 

\begin{IEEEkeywords}
    Ultrasonic SLAM, Spaitial-temporal Encoding, Generative SLAM, End-to-End 
\end{IEEEkeywords}

\section{Introduction}

SLAM (Simultaneous Localization and Mapping) technology has gained significant attention and been widely adopted across various fields. SLAM methods often use sensors such as cameras, Light Detection and Ranging (LiDAR), and millimeter-wave radar to scan unknown environments. The key features extracted from the environmental scan are used to estimate the robot's position and orientation, generating odometry, which is subsequently used to create a map of the environment \cite{yang2022sensors} .

However, performing SLAM in low-visibility conditions, such as in the presence of smoke, dust or transparent objects, presents significant challenges \cite{brunner2013selective,fritsche2017fusion,zhou2024lidar}. In such environments, the degradation of sensor performance directly affects the SLAM system's ability to gather stable and reliable environmental data. Visual sensors like cameras experience a considerable drop in performance due to severely degraded image quality, making it difficult to extract distinct features. Some image enhancement algorithms have been proposed to tackle the chanllange \cite{cho2017visibility,cho2018channel}. While LiDAR is unaffected by lighting conditions, its laser beams can scatter and attenuate in smoke and dust, resulting in incomplete or noisy point cloud data that fails to accurately capture environmental information \cite{phillips2017dust,fritsche2016radar}. Millimeter-wave radar, though capable of penetrating smoke and dust to some extent, offers lower resolution, which may not provide sufficient environmental detail \cite{yang2022sensors}. Traditional SLAM methods depend on precise feature matching and continuous pose estimation, but in adverse environments, the rise in data noise and uncertainty makes it difficult for the system to maintain accurate localization and map construction. It is common for SLAM algorithms to integrate data from multiple sensors using either tight or loose fusion methods to enhance precision \cite{fritsche2016radar,fritsche2017fusion}. While this approach requires extensive calibration and computational resources, sensor fusion remains a promising technique that has demonstrated significant improvements in SLAM performance \cite{yu2022multi,song2024learning}.

In such low-visibility conditions, ultrasonic ranging technology shows certain advantages. Since ultrasonic waves are mechanical waves, their propagation is less affected by smoke, dust, and other particulates, allowing for relatively stable distance measurements in these environments \cite{zhang2018research}. However, despite these advantages, ultrasonic sensors are rarely used in SLAM systems, primarily due to inherent limitations such as low measurement accuracy, poor angular resolution, and slow data update rates \cite{cadena2016past,ullah2024mobile}. These shortcomings make it difficult for ultrasonic sensors to perform high-precision SLAM tasks. In some studies, methods utilizing error models and sensor arrays have been employed to obtain object contours, demonstrating successful outcomes \cite{ouabi2023pose,yuan2019dsmt}. Some studies have attempted to use multiple sensors installed in a ring and achieve high-precision scanning by parsing sensor data \cite{kim2010high,kim2010optimally}. Typically, ultrasonic sensors are employed as supplementary tools to other sensor-based SLAM methods, enhancing the system's robustness and reliability in specific environments \cite{brunner2013selective,tran2021environment,ahn2008practical}.

\begin{figure}[ht]
    \centering
    \includegraphics{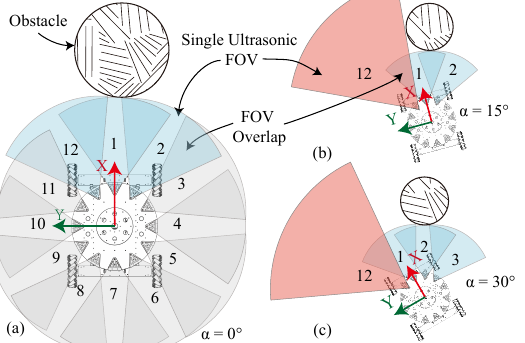}
    \caption{Angular resolution enhancement process, for example at a specific distance. The robot's sensor aligned with the x-axis is designated as sensor $1$, with subsequent sensors numbered in a clockwise direction, culminating in sensor $10$, which aligns with the y-axis. As the robot rotates, sensors $12$, $1$, and $2$ are the first to detect obstacles. The overlapping and non-overlapping detection areas of sensor $1$, along with sensors $12$ and $2$, encompass the obstacle, while the other two sensors only cover the overlapping regions. During the robot's rotation, sensor $12$ is the first to disengage from the obstacle, followed by the engagement of sensor $3$. Throughout this process, the overlapping and non-overlapping areas of each sensor gradually sweep across the obstacle, revealing that spatial features are embedded within the detection results. If sensor data is processed continuously over time, these hidden spatial features can be reconstructed.}
    \label{fig:spatial}
\end{figure}

In this work, we propose an End-to-end Generative Spatial-Temporal Ultrasonic Odometry and Mapping (EGST-UOAM) framework. We leverage the low angular resolution of ultrasonic sensors by arranging multiple sensors evenly construting an array, ensuring that their fields of view (FOV) have more than 50\% overlap. As the robot moves, the overlapping and non-overlapping regions gradually scan obstacles, generating low-resolution scan data that encode obstacle feature information implicitly. Figure \ref{fig:spatial} illustrates the enhancement of angular resolution as an example. The characteristic of ultrasonic sensors is that they will return the detection distance of the nearest obstacle within the FOV, and will not continue to detect other obstacles farther away. This is because ultrasonic sensors tend to only detect the first and strongest echo. At the current detection range shown in Fig. \ref{fig:spatial}, the sensor array can respond to approximately $15$-degree rotations of the robot, significantly improving angular resolution compared to the sensors' native lower resolution ($65$-degree). This resolution varies with the distance between the robot and obstacles, providing dynamic adaptability in different scenarios. 

To process this data, we create a sliding window for the current time frame and several preceding ones, encoding continuous temporal information. The combined spatial and temporal data in sliding window is then enhanced into a dense 2-D scan point cloud through a transformer module. A convolutional neural network (CNN) module subsequently processes the spatiotemporal feature data along with multi-frame scans to estimate the robot's pose transformation. By linearly combining the continuous scan data with the pose transformation information, both the odometry and map can be generated.

Our contributions can be summarized as follows: (i) We proposed a spatiotemporal coding method that combines a novel ultrasonic sensor array, robot movement, and a data sliding window to achieve spatial and temporal encoding of obstacle information, significantly enhancing the sensor array's obstacle characterization capabilities. (ii) We implement an end-to-end generative ultrasonic SLAM framework that addresses the inherent low-resolution limitations of ultrasonic sensors, enabling real-time updates of both maps and odometry at the sensor's frequency. (iii) Several real-world experiments validate the feasibility and robustness of the proposed framework.

\section{Proposed Method}

\begin{figure*}[ht]
    \centering
    \includegraphics{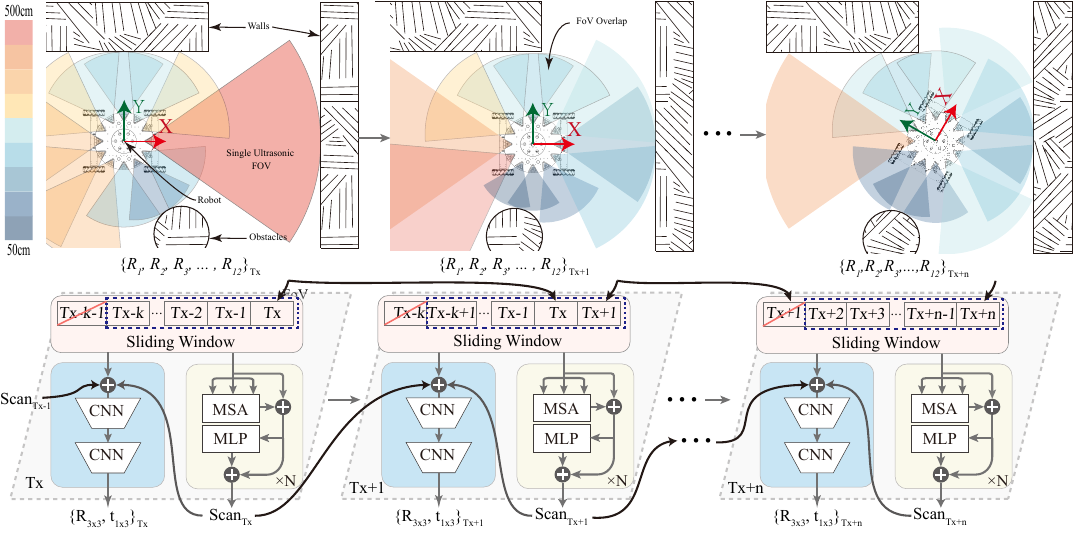}
    \caption{System overview of proposed EGST-UOAM framework at multiple timestamps. The figure illustrates how the robot spatially encodes in continuous states and how the proposed framework temporally encodes and processes the spatially encoded data.  At time $T_x$, the robot is on a random position $(a)$, and the data captured by the ultrasonic sensor array forms a $1 \times 12$ vector. As the robot moves to position $(b)$ and reaches to position $(c)$ at several time frames later, the overlapping and non-overlapping detection areas of each sensor in the array gradually sweep over obstacles, encoding spatial features into the ultrasonic detection data. The colors represent the distances of obstacles detected by individual sensors in the array. The ultrasonic sensors detect only the closest obstacles within their FOV. At each timestamp, the proposed framework first incorporates the ultrasonic sensor data from the current timestamp into the sliding window for time encoding. The sliding window data then passes through the transformer module for scan point cloud enhancement. The enhanced point cloud, along with the scan from the previous timestamp and the sliding window data, is subsequently fed into the CNN module to estimate the robot's transformation relative to its previous position.}
%  The ultrasonic sensor array, consisting of 12 sensors with 30-degree angular intervals, is mounted on a plane parallel to the X-Y plane. Each sensor has a FoV of approximately 65 degrees and a detection range between 50 cm and 500 cm. As the robot moves, the sensors' FoV and overlapping areas gradually sweep across the contours and edges of obstacles.
    \label{fig:overview}
\end{figure*}

\subsection{Overview of Proposed Method}
% The proposed framework applies sparse sensor data from an ultrasonic sensor array to better extract the latent encoded information in the signals.
The proposed EGST-UOAM framework is shown in Fig. \ref{fig:overview}.  The input to the framework consists of 12 sensors data from the current timestamp with shape of $1 \times 12$. Due to the slow speed of ultrasound propagation, changes in detection distance significantly affect the update frequency of the sensors. Therefore, we configure the sensor array to update with the latest data at each transmission point, rather than accumulating data, to ensure real-time performance. The sensor data is then fed into a sliding window within the framework, with a window size of $K$, packaging the data into a $K \times 12$ matrix.

For scan point cloud enhancement task, we adopt a weight-sharing transformer architecture, using a prediction module composed of multi-head attention mechanisms and multilayer perceptrons (MLPs) to predict scan point cloud in shape of $1 \times 720$. For the odometry task, we decompose it into the calculation of the displacement ($1 \times 3$) and rotation ($3 \times 3$) of the robot between two consecutive timestamps. The scan prediction results from the current and previous timestamps, along with the sliding window data, are fed into a CNN to estimate the robot's motion. By linearly combining the scans and transformations across consecutive timestamps, localization and mapping can be achieved.

% More specifically, the proposed EGST-UOAM framework can be represented as a tuple $Q = (S_k, F_c, F_t)$, where $S_k$ is the sliding window that concatenates the input sensor data into a matrix with temporal features, and $F_c$ and $F_t$ are the transformer and CNN networks used for the scan point cloud enhancement task and transformation estimation task, respectively. The current input sensor array data is denoted as $r$, and the temporal sliding window of the sensor array at the current timestamp, containing $K$ items, is denoted as $R_{tx}$, where $T_x$ represents the timestamp. The implicit encoded information obtained from the weight-sharing transformer is represented as $M$, and the enhanced scan point cloud is represented as $S$. This phase can be expressed as: $R_{tx} = S_k(r),\{M, S\}_{tx} = F_c(R_{tx})$

% Subsequently, the CNN module processes the sliding window data, along with the encoded information \( M \) and scan \( S \) from two consecutive frames, to estimate the robot's rotation \( R \) (represented as a \( 3 \times 3 \) rotation matrix) and displacement \( t \) (represented as a \( 1 \times 3 \) displacement vector). This phase can be represented as: $R(3 \times 3), t(1 \times 3) = F_t(M_{tx}, S_{tx}, R_{tx})$.
% Finally, the odometry \( \text{Odom} \) and map \( \text{Map} \) can be represented as: $\text{Odom} = \sum (R(3 \times 3), t(1 \times 3)),\text{Map} = \sum (\text{Odom}, S)$.

\subsection{Sensor Array Construction}
In this work, we utilize single-echo ultrasonic sensors with a FoV of approximately $65$ degrees and a detection range of $50 cm$ to $500 cm$. A total of $12$ ultrasonic sensors are mounted on the robot platform at $30$-degree intervals.  The FoV of adjacent sensors overlaps by about $30$ degrees, providing a comprehensive coverage area. The circular pattern of sensor array has been approved for encoding instantaneous spatial characteristics\cite{kim2010optimally,kim2010high}. As the robot moves, overlapping and non-overlapping parts of the sensor field of view gradually sweep over obstacles, capturing key spatial information, and temporal information is coupled in the process. This complicates the direct extraction of information from the sensor data. To address this issue, we use a transformer module to extract hidden spatiotemporal encoding features from the sensor array, thereby achieving more accurate interpretation of the environment and improving overall system performance.

% The sensors are arranged in a circular pattern, which adds complexity to the spatial encoding due to the interdependent nature of the sensors' coverage areas. The complementary relationship between sensors, combined with the robot's actual distance from obstacles, influences the encoding process. These factors result in a highly intricate spatial encoding system, making it challenging to directly extract useful information from the raw sensor data. To address this, we employ transformer module to extract the hidden spatial-temporal encoding characteristics from the sensor array, enabling more accurate interpretation of the environment and enhancing the system's overall performance.

% In this work, we arrange the sensors in a circular pattern, leveraging their spatial encoding characteristics . By coupling these spatial characteristics with temporal dynamics, we create spatiotemporal characteristics of the scan pattern, which complicates the direct extraction of information from sensor data. To address this, we employ a transformer module to extract hidden spatiotemporal encoding features from the sensor array, enabling more accurate environmental interpretation and enhancing the overall system performance.

\subsection{Scan Generation}
Our system employs a transformer module, which is composed of several submodules consist of Multi-Head Self-Attention (MHSA) block and MLP block for data generation. The input to this module is derived from a sliding window with size of K. Each block within the sliding window has dimensions of $1 \times 12$, resulting in an input size of $K \times 12$. The transformer outputs scan data with dimensions of $1 \times 720$, a resolution chosen to match the point cloud density of high-resolution 2D LiDAR, facilitating the generation of detailed maps in later stages.

As the robot moves, new data is continuously fed into the sliding window at each timestamp. The window updates by removing the oldest frame ($1 \times 12$) and appending the latest data. The transformer then processes the updated window to generate a new frame of scan data, ensuring that the system keeps up with the robot's movement. To improve the robustness of the scan generation task, we define the loss function as the sum of two terms: (1) the mean squared error (MSE) between the distance of each point in polar coordinates and the corresponding label, and (2) the MSE between the curvature of triangles formed by neighboring points $N$ and that of the label.

Due to the physical constraints of the robot, such as maximum velocity and acceleration limits, the range of motion between consecutive timestamps is restricted. This constraint allows the sliding window mechanism to ensure that the consecutive frames from the ultrasonic sensors capture spatial patterns with temporal consistency. As a result, the system faithfully records the overlap and non-overlap areas in the field of view, allowing for continuous scanning of obstacles within these regions. This design ensures that the data has a strong temporal and spatial coherence, enabling more accurate map generation.

\subsection{Transformation Estimation}
Due to the lack of smoothness in the scan point cloud generated by the scan generation task, it becomes challenging to calculate inter-frame matching using conventional methods such as ICP (Iterative Closest Point) or NDT (Normal Distributions Transform). A CNN module is employed to calculate the robot's transformation—encompassing both translation and rotation—based on the sliding window data and the generated scan point cloud from two consecutive frames. The CNN module receives three inputs: the sliding window data, the scan point clouds from the previous frame, and the current frame. The generated scan point clouds are provided by the scan generation.
% The sliding window contains raw ultrasonic sensor data from the current timestamp as well as the preceding K frames, embedding both spatial and temporal encoding relationships. 
By processing this combination of inputs, the CNN extracts the transformation information, allowing the system to accurately estimate the robot's motion between frames. This integration of spatio-temporal data from the sliding window and high-density scan data ensures that both the temporal consistency and the fine-grained spatial features of the environment are preserved, leading to precise calculation of the robot’s movement and rotation.

\subsection{Dataset Composition}
We construct a dataset for training and evaluating the proposed framework in a controlled environment, using LiDAR data as the ground truth for scans and a motion capture system as the ground-truth odometry. The scan data is captured at $10 Hz$, with each round consisting of $720$ distance measurements covering a 360-degree field of view. The motion capture system provides odometry information at a $50 Hz$ update rate. By calculating the difference between the position and orientation of the robot in continuous time from the motion capture system and performing pre-integration, a transformation relationship can be obtained at the same update frequency of the proposed framework.

\section{Experiments Result}
To verify the feasibility and robustness of the proposed framework, we conducted experiments in the real world. The sliding window $K$ is set to 3, and the curvature neighbor value of the loss function of the scan generation task $N$ is set to 1. Since ultrasonic SLAM has been rarely studied, we use 2-D lidar as a baseline for evaluation to demonstrate the effectiveness of the proposed framework.

\subsection{Experiments Setup}
\begin{figure}
    \centering
    \includegraphics{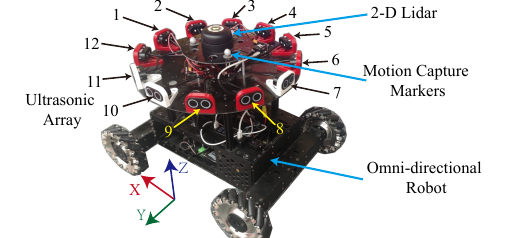}
    \caption{Robot setup used in the experiment. The onboard PC uses an AMD 8945HS CPU for framework inference. We used an omnidirectional mobile robot platform, 2-D Lidar to provide reliable scan point clouds, and a Nokov motion capture system to provide odometer annotations.}
    \label{fig:robot}
\end{figure}

The experiments were conducted using the Robomaster AI robot, which is equipped with Mecanum wheels, enabling omnidirectional movement. The ultrasonic sensors used are A21 single-echo type, with an update frequency set at 3 Hz. For ground truth data, we utilized an RPLIDAR A2 to provide scan labels and the Nokov motion capture system to supply odometry labels. The operational cycle of the framework is also set at 3 Hz, aligning with the sensor update frequency to ensure synchronized data processing and real-time performance. The detailed robot setup is shown in Fig \ref{fig:robot}.

\subsection{Result}
\begin{figure}[h]
    \centering
    \includegraphics{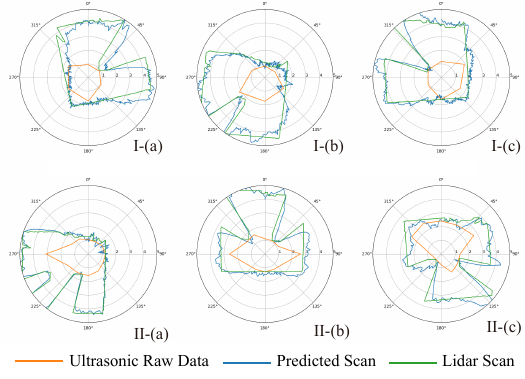}
    \caption{Comparison of the scan quality of the proposed EGST-UOAM framework and lidar. Series I indicates that there is only one obstacle in the scene, and series II indicates two arbitrary obstacles in the scene. Orange represents the raw ultrasonic array data at current timestamp, blue is the predicted scan point cloud, and green is the laser result. From the results, it can be seen that the proposed framework better restores features such as walls and obstacle positions.}
    \label{fig:scan}
\end{figure}

We evaluated the performance of the proposed EGST-UOAM framework from two aspects: mapping quality and odometry accuracy. We constructed an empty room with some obstacles inside, and the walls and obstacles of the room were made of low laser-intensity foam bricks. Figure \ref{fig:scan} shows the scan quality of the proposed framework and the comparison with the lidar method. Table \ref{tab:result} shows the mapping overlap between the proposed framework and the 2-D LiDAR Cartographer algorithm \cite{hess2016real}, as well as the odometry error. The experimental results demonstrate the feasibility of the proposed framework and achieve ultrasonic slam with the same frequency update as the sensor. 
Although the scan of the proposed framework is not as smooth as that of LiDAR, it is less affected, for example, it is able to identify walls instead of considering them as open areas as shown in Fig\ref{fig:scan}-I-(a) and II-(b).

\begin{table}[h!]
    \label{tab:result}

    \begin{tabular}{|l|l|l|l|}
    \hline
                         & Ground Truth & Lidar Cartographer & EGST-UOAM \\ \hline
    Map Overlap          &        100\%      &        91.7\%           &      73.2\%      \\ 
    Odometry             &         100\%     &       87.1\%           &          69.5\%   \\ 
    Map Rate      &      N/A        &        1Hz            &      1Hz     \\ 
    Odometry Rate &        N/A      &        5Hz            &      3Hz     \\ \hline
    \end{tabular}
    \end{table}

\section{Conclusion}
In this work, we propose the EGST-UOAM framework, which integrates a novel ultrasonic sensor array structure with robot motion coupling for spatial obstacle encoding. Temporal encoding is accomplished using a sliding window mechanism. The spatiotemporally encoded data is then leveraged to generate dense scan point clouds and robot pose transformations for map construction and odometry. This framework overcomes the inherent limitations of ultrasonic sensors, enabling SLAM with high accuracy and real-time updates at the sensor's frequency. Several real-world experiments validate the feasibility and robustness of the proposed approach, demonstrating its enhanced obstacle representation capabilities and ability to achieve real-time updates of both maps and odometry.

\bibliography{icassp2025}
\bibliographystyle{IEEEtran}

\end{document}